# QMUL_AK at AISA-ArabicFC: Schema-Constrained Function Calling with ALLaM

**Ali Khairallah** **Arkaitz Zubiaga**
Queen Mary University of London
a.m.khairallah@qmul.ac.uk a.zubiaga@qmul.ac.uk

## Abstract

We describe the QMUL_AK submission to AISA-ArabicFC 2026. The system adapts the Arabic-English ALLaM-7B-Instruct model to function calling with low-rank adaptation (LoRA). It combines an Arabic schema-aware prompt, dialect and no-call resampling, a token-weighted training objective, and deterministic validation of generated calls against the candidate tool schema. The same output supports Track A function calling and Track B reasoning-augmented calling. Training uses only the official task data. On the development set, the system achieves 0.9982 function accuracy and 0.8920 argument exact match, corresponding to overall A and B scores of 0.9345 and 0.9454. On the blind test set, the system reaches overall A and B scores of 0.9253 and 0.9380, ranking 3rd on both submission tracks and 1st of 21 teams on the Track C dialect-robustness diagnostic with a macro-averaged score of 0.9075 and the smallest dialect gap in the shared task (0.077). The results show that function selection is nearly saturated, while argument extraction remains the main source of error.

## 1 Introduction

AISA-ArabicFC evaluates whether a system can decide when an Arabic request requires a tool, select a function from a supplied candidate set, and return its arguments as structured JSON (Nacar et al., 2026a). It covers Arabic in Modern Standard Arabic (MSA) and four regional dialect groups across practical service domains. The task is connected to the AISA reference architecture (Nacar et al., 2026c) and to a recent data-centric work on reliable Arabic function calling (Nacar et al., 2026b).

Our submission uses ALLaM-7B-Instruct, an Arabic-English autoregressive model (Bari et al., 2025), rather than a larger general-purpose one. We fine-tune it with LoRA (Hu et al., 2022) and keep the base weights in `bfloat16`. Three design choices address the main difficulties of the task: an Arabic schema-aware prompt that places the candidate functions, their argument names, and types directly in the input; a resampling scheme that offsets the imbalance of no-call and minority-dialect examples; and a deterministic validation step that removes malformed or out-of-schema fields before submission. The rest of this paper describes the methodology implemented by our system and analyses the results achieved, which placed third on both submission tracks and first on the dialect-robustness diagnostic.

Code, the submitted adapter, and reproduction instructions are available at https://github.com/alikhairallah/AISA-ArabicFC.

## 2 Data

The shared task provides 10,550 training, 545 development, and 1,125 blind test examples (Nacar et al., 2026a). Each example contains a user request in MSA or one of four regional dialect groups (Gulf, Levantine, Egyptian, and Maghrebi), a developer message carrying date and time, and a candidate set of four tool schemas drawn from a registry of 27 functions. Each schema declares a function name, an Arabic description and typed argument definitions. Of the 27 registry functions, 20 appear as gold calls; the remaining seven occur only as distractors in the candidate sets and are never correct. The reference output is a single call naming one of the 20 callable functions together with an argument dictionary or `none` when no tool applies. Arguments are heterogeneous, with identifiers and free-text spans copied from the request, numeric quantities, closed-class codes such as ISO currency and language identifiers, and comma-separated lists. Table 1 shows one example with its reference call. No-call examples are rare in training (50 of 10,550, 0.5%) but frequent in development (45 of 545, 8.3%), which motivates the resampling scheme in Section 3. Every example additionally carries a dialect label, which we use for the analysis in Section 5.

| | |
|---|---|
| Request (AR) | الجو في الدوحة بعد ثلاث أيام؟ |
| Translation | "The weather in Doha in three days?" |
| Gold | `{"tool_called": "get_weather", "arguments": {"city": "الدوحة", "days": 3.0}}` |

Table 1: A training example. The city الدوحة ("Doha") is copied verbatim into the argument, while ثلاث ("three") is written as a word in the request and must be emitted as the number 3.0.

## 3 System

**Schema-aware prompting.** Each example contains a user request, a developer instruction, and a set of candidate tools. We convert every sample into a compact Arabic description containing the function name, description, argument names, types, and argument descriptions. These descriptions are appended to the original developer message. A final Arabic instruction requires a short <think> segment followed by one JSON object with the keys `tool_called` and `arguments`. It also states that only information explicitly present in the request may be extracted. No-call cases must use `none` with an empty argument object. The resulting system and user messages are rendered with ALLaM's native chat template.

This representation makes the output contract part of every training and inference instance. It also avoids maintaining a fixed global tool inventory: the model sees only the functions that are supplied for that row.

**Model adaptation.** We fine-tune ALLaM using LoRA modules applied to the query, key, value, and output attention projections, as well as the gate and feed-forward projections in each transformer block. The pretrained model remains frozen while only the LoRA parameters are optimised.

**Training objectives.** Training uses token-level causal cross-entropy over the generated completion. Prompt tokens are excluded from the loss, while JSON function-call tokens receive a weight of 1.0 and reasoning tokens inside the <think> segment receive a lower weight of 0.3. This assigns greater importance to the exact function name and argument structure while retaining the reasoning format required for Track B.

**Data balancing.** The training split is heavily imbalanced: only 50 of the 10,550 examples are no-call, and the dialects are unevenly represented. We correct both by replicating no-call examples and by duplicating every example according to a dialect weight (3 for Maghrebi, 2 for Levantine and Egyptian, and 1 for Gulf and MSA), so scarce cases are seen more often. For the submitted model we combine the train and development splits (11,095 examples) and resample the no-call examples to match the no-call share produced by the development-phase resampling, approximately 11%, which keeps the total divisible by the batch size. This gives 16,308 training samples. Development-phase configuration applies the resampling to the training split only, totalling 15,583 examples, of which 1,740 are no-call. Table 2 shows the distribution before and after resampling.

| | Train | Dev | Test | Train∪Dev resampled |
|---|---|---|---|---|
| MSA | 6,151 | 323 | 825 | 7,550 |
| Levantine | 1,784 | 75 | 102 | 4,150 |
| Egyptian | 1,283 | 66 | 104 | 2,759 |
| Gulf | 1,190 | 75 | 83 | 1,405 |
| Maghrebi | 142 | 6 | 11 | 444 |
| Total | 10,550 | 545 | 1,125 | 16,308 |
| no-call | 50 | 45 | — | 1,820 |

Table 2: Dataset composition by dialect. The final column is the training distribution of the submitted model, obtained by resampling train ∪ dev (11,095 examples) to 16,308. No-call is a subset of each column; test gold was not released. No-call prevalence shifts from 0.5% in train to 8.3% in dev, which motivates the resampling in Section 3.

### 3.1 Output validation

We extract the first valid JSON object after the optional reasoning segment and validate it against the candidate functions provided for the input. Predictions that select an unavailable function are mapped to a no-call output. For a valid function, we retain only arguments declared in its schema and remove null, empty, non-finite, and out-of-enumeration values. Values are not checked against the declared primitive type. This step enforces the required output format without selecting another function or constructing new argument values.

After validation, we replace the model's free-form reasoning with a short deterministic Arabic statement associated with the validated function, including a corresponding statement for no-call predictions. This ensures that the submitted trace is non-empty and consistent with the final action, but it also means ThinkRate is satisfied by construction rather than by model reasoning: the reported Track B scores measure trace presence, not faithfulness. The procedure does not change the predicted function or its arguments.

## 4 Experimental setup

**Data protocol.** Development experiments use the official training split for adaptation and the 545-row development split for evaluation. All development results and ablations reported here come from models trained on the official training split alone and evaluated on the held-out split. During the blind test, the produced checkpoint was retrained separately on the union of train and development, so the development figures are not held-out measurements of the checkpoint scored on the blind test set. After fixing the prompt, sampling strategy, and training configuration, we retrained the submitted model on the union of the training and development splits and generated predictions for the 1,125 blind test examples. The test split is not used for model or configuration selection. We use no external function-calling data, test labels, or labelled resources that overlap with the task data. The organisers verified top submissions for reproducibility and fairness. Gold test labels were not released, so all reported test-set scores are the official leaderboard scores, including the per-dialect breakdown published for Track C.

**Evaluation.** We report function accuracy (FnAcc), argument exact match over positive examples (ArgEM), and reasoning presence rate (ThinkRate). The official track scores, denoted A and B, are computed as:

$$A = 0.4\,\mathrm{FnAcc} + 0.6\,\mathrm{ArgEM},$$
$$B = 0.3\,\mathrm{FnAcc} + 0.5\,\mathrm{ArgEM} + 0.2\,\mathrm{ThinkRate}.$$

**Implementation.** Table 3 lists the full training and inference configuration. We retain ALLaM frozen base parameters in `bfloat16`, while the trainable LoRA parameters remain in `float32` (no quantization is applied). Experiments use PyTorch 2.11, Transformers 4.46.3, PEFT 0.19.1, and torchao 0.17.0 on a single-GPU Google Colab G4 runtime. The system does not use torchao quantization; the version is pinned for reproducibility purposes, as PEFT's optional torchao availability probe is incompatible with the package preinstalled on Colab. Training the final model for two epochs takes approximately 65–75 minutes.

| Setting | Value |
|---|---|
| LoRA rank / $\alpha$ / dropout | 16 / 32 / 0.05 |
| Epochs / batch size | 2 / 12 |
| No-call oversampling factor | 30 |
| Maximum sequence length | 2,048 tokens |
| Optimizer / learning rate | AdamW / $10^{-4}$ |
| Learning-rate schedule | Cosine, 5% warm-up |
| Gradient clipping | 1.0 |
| Decoding | Greedy, 320 new tokens |
| Random seed | 42 |

Table 3: Training and inference configuration of the submitted system.

## 5 Results and Analysis

**Overall results.** Table 4 reports development and blind-test scores. The system predicts the correct function for 544 of the 545 development examples. Argument objects are exactly correct for 446 of the 500 positive examples, leaving 54 argument errors but only one function error. The main performance bottleneck is therefore argument extraction rather than deciding whether to call a tool or selecting the appropriate function. Since ArgEM is evaluated over the complete argument object, a single missing, additional, or incorrectly normalised value makes the entire prediction incorrect. All results in this paper come from single runs at seed 42; we do not report variance across seeds.

| Split | FnAcc | ArgEM | Overall A | Overall B |
|---|---|---|---|---|
| Development | 0.9982 | 0.8920 | 0.9345 | 0.9454 |
| Test | 0.9938 | 0.8796 | 0.9253 | 0.9380 |

Table 4: Development and official blind-test results, computed with the official scorer. ThinkRate is 1.0000 on both splits. QMUL_AK ranked 3rd on both submission tracks.

**Dialect breakdown.** Argument extraction on the development set also varies across dialect. The system obtains an ArgEM of 0.9273 for MSA (268/289), 0.8889 for Gulf (64/72), 0.8485 for Egyptian (56/66), 0.8060 for Levantine (54/67), and 0.6667 for Maghrebi (4/6). The Maghrebi result is based on only six positive examples and is therefore not a stable estimate. Nevertheless, the gap between MSA and the larger Egyptian and Levantine subsets indicates that robust dialectal argument extraction remains an open challenge.

**Generalisation from development to test.** Our Overall A score falls by 0.009 between development (0.9345) and blind test (0.9253), and Overall B by 0.007 (0.9454 to 0.9380). Function accuracy falls by 0.004 and argument exact match by 0.012, so argument extraction remains the bottleneck on unseen data.

**Dialect robustness on the test set.** Track C macro-averages the per-dialect overall score with equal weight per dialect and reports the gap between a system's best and worst dialect. Our system ranks 1st of 21 teams with a macro score of 0.9075, scoring 0.935 on MSA, 0.932 Egyptian, 0.915 Gulf, 0.897 Levantine, and 0.859 Maghrebi. Both the Maghrebi score and the gap of 0.077 are the best recorded in the shared task. An ablation on the development split (Table 5) shows that removing dialect resampling lowers overall ArgEM by 1.0 point but raises Maghrebi from 0.6667 to 0.8333 and Levantine from 0.8060 to 0.8358. With six Maghrebi positives in development these are one- and two-example differences, so the development split cannot resolve the per-dialect contribution either way.

**Component ablation.** Table 5 removes one component at a time, training on the training split alone and evaluating on the held-out development split. Every removal costs accuracy, so the full configuration is the best of the four. Token weighting is the largest lever, costing 2.0 ArgEM points when reasoning tokens are given full weight, and it is the only controlled comparison in the table: it leaves the training frame unchanged at 15,583 rows and 2,596 optimiser updates. Removing dialect weights costs 1.0 ArgEM, and removing no-call oversampling costs 0.6, but both also shrink the training frame, so part of each gap may reflect fewer updates rather than the component itself.

**Contribution of the validation layer.** Table 6 compares the model's parsed output with the validated output over the same generations. Validation changes 1 of the 545 development rows and removes a single argument field. It raises FnAcc from 0.9963 to 0.9982 and leaves ArgEM unchanged at 0.8920, worth 0.0007 Overall A. The model selects a function outside the candidate set on one row, which is mapped to none; the gold for that row is also none, and no row with a gold call is ever mapped away, so the mapping discarded nothing recoverable. There are no parse failures.

| Configuration | Rows | FnAcc | ArgEM | Overall B |
|---|---|---|---|---|
| Full system | 15,583 | 0.9982 | 0.8920 | 0.9454 |
| – dialect weights | 12,000 | 0.9890 | 0.8820 | 0.9377 |
| – no-call $\times 30$ | 13,901 | 0.9853 | 0.8860 | 0.9386 |
| – token weighting | 15,583 | 0.9963 | 0.8720 | 0.9349 |

Table 5: Component removal on the held-out development split, training on the training split only at seed 42. Removing a resampling component shrinks the training frame and so the number of optimiser updates (2,596, 2,000 and 2,316 respectively); only the token-weighting row is directly comparable to the full system. Token weighting is the largest single contributor to ArgEM.

| Output | FnAcc | ArgEM | Overall A | Overall B |
|---|---|---|---|---|
| Parsed only | 0.9963 | 0.8920 | 0.9337 | 0.9449 |
| Validated | 0.9982 | 0.8920 | 0.9345 | 0.9454 |

Table 6: Effect of validation on the development split, scored over the same generations with an identical deterministic trace.

## 6 Conclusion

In this paper, we present the QMUL_AK system for Arabic function calling we implemented as part of the AISA-ArabicFC shared task, based on LoRA adaptation of ALLaM with schema-aware prompting, imbalance-aware sampling, and constrained output validation. Development results show that function selection is highly reliable, while exact argument extraction accounts for almost all remaining errors. In the test phase we ranked 3rd on Tracks A and B and 1st of 21 teams on the Track C dialect-robustness diagnostic. Future work should therefore focus on dialect-robust value extraction and normalisation methods that generalise beyond development-specific patterns.

## Limitations

All results come from single runs at seed 42, so component effects cannot be separated from run-to-run variance. Removing a resampling component also changes the number of optimiser updates, so only the token-weighting ablation is a controlled comparison. We evaluate on one tool registry and do not test transfer.

## Acknowledgements

We thank Tuwaiq Academy and the AISA-ArabicFC organizers for providing the dataset, the baseline model, and the evaluation framework for this shared task.